# Model-Based Compensation of Moving Tissue for State Recognition in Robotic-Assisted Pedicle Drilling

Zhongliang Jiang, Long Lei, Yu Sun, Xiaozhi Qi, Ying Hu,
Bing Li, Nassir Navab, and Jianwei Zhang

*Abstract*— **Drilling is one of the hardest parts of pedicle screw fixation, and it is one of the most dangerous operations because inaccurate screw placement would injury vital tissues, particularly when the vertebra is not stationary. Here we demonstrate the drilling state recognition method for moving tissue by compensating the displacement based on a simplified motion predication model of a vertebra with respect to the tidal volume. To adapt it to different patients, the prediction model was built based on the physiological data recorded from subjects themselves. In addition, the spindle speed of the drilling tool was investigated to find a suitable speed for the robotic-assisted system. To ensure patient safety, a monitoring system was built based on the thrusting force and tracked position information. Finally, experiments were carried out on a fresh porcine lamellar bone fixed on a 3-PRS parallel robot used to simulate the vertebra displacement. The success rate of the robotic-assisted drilling procedure reached 95% when the moving bone was compensated.**

*Index Terms*—**Surgical robotics, medical robotics, robot-assisted spinal surgery; surgical state recognition; parameter identification; vertebra motion model.**

## I. Introduction

THE spine ages, due to disc degeneration, resulting spinal lesions often occur with instabilities and deformities. Pedicle screw fixation is one of the most popular treatments for patients with spinal instability and deformity [1, 2]. To achieve the best biomechanical stability, the desired screw path is ended inside of the inner cortical layer, as shown in Fig. 1. However, the pedicle drilling procedure is one of the most dangerous operations in orthopedic surgery because vertebra is surrounded by many vital tissues, such as nerves and blood vessels.

This research supported by the National Natural Science Foundation of China (Nos. U1713221 and U1713218), Shenzhen Fundamental Research Funds (No. JCYJ20180507182215361). (*Zhongliang Jiang and Long Lei contributed equally to this work.*) (*Corresponding author: Ying Hu.*)

Zhongliang Jiang is with Computer Aided Medical Procedures, Technische Universität München, 85748 Garching, Germany, and also with Shenzhen Institutes of Advanced Technology (SIAT), Chinese Academy of Sciences, 518055 Shenzhen, China (zl.jiang@tum.de).

Xiaozhi Qi and Ying Hu are with SIAT, Chinese Academy of Sciences, Shenzhen, 518055, China (ying.hu@siat.ac.cn).

Long Lei, Yu Sun and Bing Li are with Harbin Institute of Technology (Shenzhen), Shenzhen, 518055, China.

Nassir Navab is with Computer Aided Medical Procedures, Technische Universität München, 85748 Garching, Germany, and also with the Johns Hopkins University, MD, USA.

Jianwei Zhang is with University of Hamburg, D-22527 Hamburg, Germany.

The authors would like to thank Dr. Binsheng YU from Department of Spine Surgery, Peking University Shenzhen Hospital, Shenzhen, China for their valuable feedback and discussions.

In addition, the largest transverse pedicle width of thoracic vertebras is T12 (8 mm, see Fig. 1) [3, 4], and the diameters of bone screws for T12 are usually 5–5.5 mm, so that the acceptable error shouldn't be more than 1.5 mm. Furthermore, the surgeons are not able to see the tip of the drill because it is covered by sawdust and blood. Surgeons need to distinguish drilling states purely based on tactile feedback [5], and decide the next action (drilling ahead, changing the direction or stop drilling) in a short time.

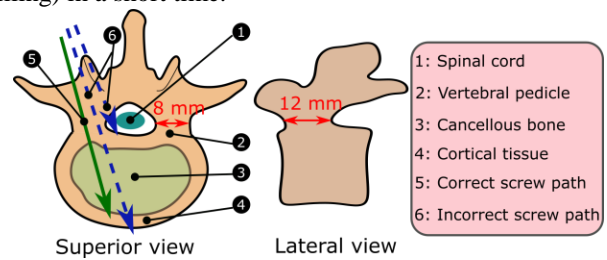

Fig. 1. Diagram of screw paths on a vertebra. The desired path is ended inside of the inner cortical bone as green line. The dashed lines are incorrect examples, which will puncture the spinal cord or thoroughly penetrate the cortical bone.

To address the challenges, Schatlo and Dijk *et al.* reported that a robotic-assisted system (RAS) called SpineAssist can improve the accuracy rate of pedicle screw placement, shorten operation time, and reduce radiation [2, 6, 7, 8]. However, although it can provide operational guidance, the drilling procedure is still manually performed.

To develop a more advanced RAS for spinal surgeries, automatic or semi-automatic RASs have been proposed [9-14]. Sun *et al.* proposed a state recognition method for bone drilling process by combining the audio and force data [15]. Lee *et al.* proposed automated surgical planning based on computed tomography (CT) images for fusion surgery [16]. Dai *et al.* used the vibration signal to identify the states of drilling and milling [17, 18]. Giovanni *et al.* built a telerobotic system with force feedback to perform pedicle drilling [19]. Sun *et al.* automatically generate grinding trajectory using 3D imaging for decompressed laminectomy [20]. Hu *et al.* defined different phases of the drilling process [21]. Dai *et al.* used a support vector machine to identify different tissues [22]. Jiang *et al.* predicted the milling depth purely based on the milling force without the requirement for any tracking system [23]. To guarantee the safety of surgery, the force information is important for reflecting the changes in drilling conditions. It has been widely used in medical applications, such as robotic ultrasounds [24].

However, most of the aforementioned automatic RASs are validated only when the tested bone is stationary. However, the vertebra is moving due to the respiration, increasing the difficulty of pedicle drilling. Moreover, recent research has shown the influence of beating heart on robotic heart surgery [25-27]. To predict the spinal movement, Bauer and Ignasiak *et al.* built a multi-body model (including ribs, vertebras and sternum) with given mechanical properties [28, 29]. The relative prosperities have been studied in [30-32]. The multi-body model is mainly used for qualitative analysis, such as crash car test and cannot be used in spinal surgery, which requires accurate prediction of vertebral motion since it is hard to accurately obtain specific mechanical properties for each patient, especially during operation.

In addition, Winklhofer *et al.* validated that the spinal motion is mainly caused by respiration [33]. The respiration system can be divided into three parts: inhaled/exhaled air, the lung, and the chest wall, which involves the spine [30]. Thus, the respiration-based chest volume changes will further result in vertebra displacement. Thus, this work directly builds a vertebra displacement model with respect to the tidal volume ($Tv$).

To detect the right stop point when the tested bone is not stationary, this paper first builds a vertebra displacement model with respect to $Tv$, then uses the prediction model for compensation, and further detects the right stop point for drilling procedure based on the force information. The parameters in the displacement model are optimized by the particle swarm optimization (PSO) algorithm based on the physiological data recorded directly from the subjects. Thus, the model with different parameters is able to adapt to different patients with various tissue properties. To integrate the method into clinical routine, $Tv$ is calculated from the ventilator's setting because general anesthesia is usually required in spinal surgery. Eventually, the force-based identification algorithm is tested on the fresh lamellar bone fixed on a parallel robot by simulating the vertebra displacements in three orthogonal directions.

This work is built on the preliminary work presented in [34]. Here we further considered integrating the state recognition method for moving tissue into the current clinical routine. The main contributions of this work are as follows:

- The shoulder and elbow joints of a self-designed robotic spinal surgery system (RSSSII) is redesigned to improve the stiffness for performing bone drilling procedure.
- A physical model is introduced to describe the recorded physiological data, rather than using an empirical model.
- A method to compute the real-time $Tv$ from the ventilator's setting used in surgery and a safety monitor system to ensure patient safety based on both drilling force and position.
- To effectively identify the drilling state using RAS, the spindle speed is investigated to select a suitable speed for RAS rather than using the value often used by surgeons.

The rest of this paper is organized as follows. Section II describes the structure of RSSSII and Spinal Physiological Motion Simulator (SPMS) used to simulate the vertebra displacement. Then, the vertebra displacement model based on a simplified spinal configuration is given in Section III. Section IV describes the compensation method with a safety monitoring system and the state recognition algorithm. The experimental results on a porcine lamellar bone are in Section V.

## II. HARDWARE DESCRIPTION

### A. Scheme of RSSSII

Compared with the previous version of the spinal surgery system [9], RSSSII is partially modified to improve the stiffness of the arm. The overall structure of RSSSII has been shown in Fig. 2 A. The 6-DOF robot arm would be used to adjust the position and orientation of the drill, and the 2-DOF operation device is used to drill the screw path. Comparing with SpineAssist [8], the serial multi-joint robot (RSSSII) has a larger working space, which makes it easier to move from one vertebra to another one. In the new version, spline joints are both used in the shoulder joint and elbow joint. The new scheme of the shoulder joint, shown in Fig. 2 B, greatly increases the effective length in axis direction, reducing the local maximum axial stress. In addition, the details of the 2-DOF operation device and electronics systems have been presented in Fig. 2 C and Fig. 2 D, respectively.

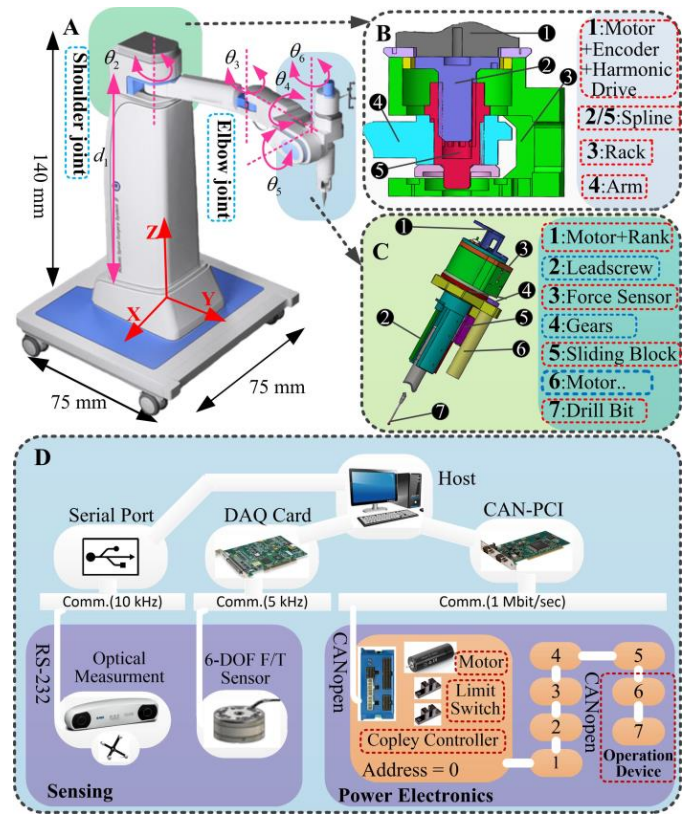

Fig. 2. Design of RSSSII. A. overall structure, B. structure of shoulder joint, C. 2-DOF operation device, and D. involved sensing and electronic system.

### B. Design of SPMS

SPMS is a parallel mechanism used to simulate the vertebra displacement in operation (see Fig. 3). The SPMS consists of the exactly same three branch chains connected to the moving platform. The details of each branch chain is depicted as follows: the motor drives the lead screw by a drive belt; the rotation of the lead screw results in slider movement along the axis of the lead screw; a revolute joint is fixed on the slider, and a fixed-length rod connects the slider and passive spherical joint, which is screwed into the moving platform. Since the parallel mechanism has good stiffness and load capacity, the accuracy of the SPMS can be guaranteed during the drilling process.

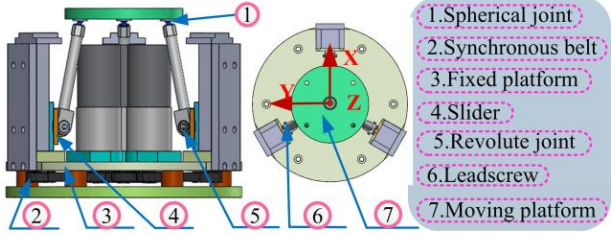

Fig. 3. Overview of the SPMS. Profile view and overhead view are shown in left and right, respectively.

### III. PREDICTION MODEL OF VERTEBRA DISPLACEMENT

The vertebra displacement was recorded from five volunteers via a NDI camera. The discrete wavelet transform (DWT) was used as a preprocessing method. A mathematical model of vertebra displacement was deduced based on the simplified spine configuration. The patient-relevant parameters were adjusted by the PSO algorithm based on processed displacement.

#### A. Physiological Signals Acquisition and Processing

All physiological signals were recorded from five healthy volunteers without an orthopedic disease, who were instructed to lie face down on the operation bed (see Fig. 4). The three orthogonal directions included anterior–posterior (AP), superior–inferior (SI), and left–right (LR). Since the T12 segment has an anatomical landmark that can be easily located by palpation [35], the vertebral displacement data were recorded by the attached passive markers on the skin above T12.

The displacement and $Tv$ were recorded by a Polaris camera (NDI, Ontario, Canada) and a Gas Flow Sensor (Siargo, Santa Clara, USA), respectively. Since the breathing rate of the healthy adults is around 20 times per minute, the sampling frequency ($Fs$) was set to 8 Hz and 64 Hz for displacement and $Tv$, respectively. All data were recorded during regular breathing (about 30 s).

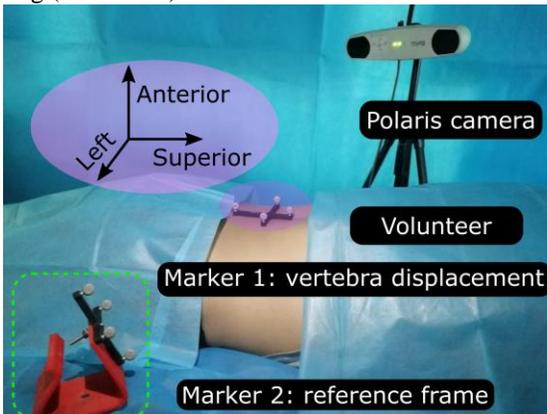

Fig. 4. Actual acquisition experiment, Marker 1 is attached to the skin of subjects, Marker 2 is used as reference coordinate system used to depict the vertebra displacement.

Since the measured displacement is disturbed by the environment noise (e.g. digital noise from the sensor, the slight movements of patients), a preprocess method is necessary to remove the noise. But the displacement is not a typical periodic signal. To obtain good resolution both in time and frequency domain, DWT was employed for de-noising. An example of DWT decomposition coefficients of the vertebra displacement along anterior-posterior direction is given in Fig. 4.

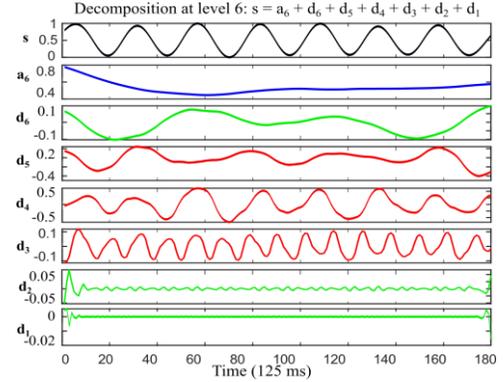

Fig. 5. DWT Decomposition of displacement. The used wavelet basis is db5, and decomposition level is 6. The original signal is denoted by **s**. The $a_6$ and ($d_1$, $d_2$, $d_3$, $d_4$, $d_5$, $d_6$) represent the approximation and detail coefficients for 6 levels.

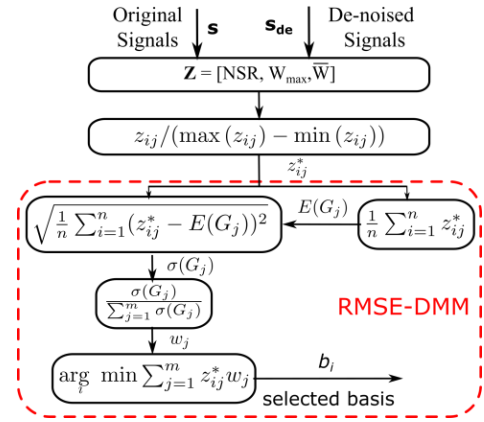

Fig. 6. Selection of suitable wavelet basis function. $z_{ij}$ is the value of $j$th metric when the $i$th base function is used in DWT.

To calculate the detail ($d_i$) and approximation ($a_i$) coefficients, $a_{i+1}$ from higher level is passed through highpass and lowpass filters, followed by downsampling by two. Thus, $d_4$ represents the frequency range [$Fs/32$, $Fs/16$] containing the dominant frequency of breathing. This is also reflected by the magnitude of $d_4$, which is fluctuated between -0.5 and 0.5, which is much larger than the magnitude of other $d_i$. To make the de-noising process adapt to different patients, the de-noised signal ($s_{de}$) is calculated as follows:

$$s_{de} = d_3 + d_4 + d_5 \qquad (1)$$

To quantitatively evaluate the de-noising performance, three metrics were used [34]. First the NSR, the reciprocal of SNR (signal-to-noise ratio), is taken into consideration. Then, the difference ($L_i$) from the smallest to the largest displacements with same $Tv$ is employed to reflect the effectiveness of the de-noising method for the data recorded in different breathing periods. To use $L_i$ for discrete data, the narrowly neighbored area around $Tv$ ($U(Tv, \Delta Tv)$) was used rather than a single $Tv$.

$$L_i = \max(s_{de}(U(Tv, \Delta Tv))) - \min(s_{de}(U(Tv, \Delta Tv))) \qquad (2)$$

Then the three evaluation metrics are defined by Eq. (3).

$$\begin{bmatrix} NSR & W_{max} & \overline{W} \end{bmatrix}^T = \begin{bmatrix} \dfrac{1}{SNR} & \max(L_i) & \dfrac{1}{N}\sum_{i=1}^{N} L_i \end{bmatrix}^T \quad (3)$$

where $W_{max}$ and $\overline{W}$ represent the maximum and average of all involved $L_i$.

To achieve the best de-noising performance for different displacement signals from different patients, the RMSE-DMM [36] is used to select the optimized basis functions, as shown in Fig. 6. The key is to reasonably allocate the corresponding weights for different metrics. Then, the best basis function is selected by looking for the weighted sum of three metrics achieving the smallest value.

*B. Model of Vertebra Displacement*

The sternum will move up and out during inspiration because of the expansion of chest volume (see Fig. 7). Due to the friction between the body and bed, the sternum's motion would be restricted while the motion of the spine would be larger in spinal surgery. Thus, the vertebra displacement is described in orthogonal directions.

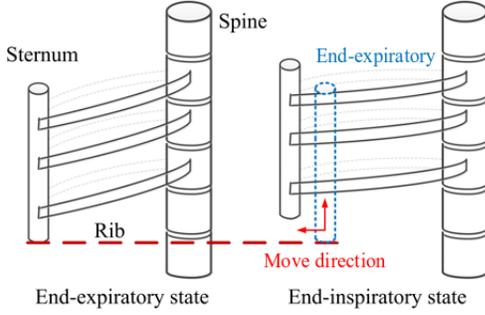

Fig. 7. Lateral view of chest. The left and right parts depict the typical chest movements caused by breathing without any additional constraints.

The spinal configuration is depicted in Fig. 8 a), consisting of a vertebra and two intervertebral discs deployed on different sides. Thus, a simplified physical model of the spinal configuration is shown in Fig. 8 b). Since vertebra displacements vary according to the changes in the intrathoracic pressure, a tensile/compressive force ($P$) acting along the vertebral axis together with a uniform load of intensity ($q$) acting throughout the span of the spine are used to depict the varying loads caused by respiration. In addition, the force in LR direction is considered to be zero because the chest is close to being systematic about the spine. Assuming that the state of the inhaled air does not change, q can be deduced according to the ideal gas equation of state as follows:

$$q = \dfrac{Tv}{V_0} p_0 b_w \quad (4)$$

where $V_0$ is the initial volume the chest, $p_0$ is the atmospheric pressure, $b_w$ is a constant length along the LR direction.

$$P = \dfrac{Tv}{V_0} p_0 S \quad (5)$$

where $S$ represents the equivalent area of the chest slices along AP direction. Since the change of the chest size is small in the plane perpendicular to the AP direction, S is seen as constant.

Then, the length of vertebras (e.g. T12) and the two neighboring discs ($l$, $m_1$, $m_2$) were measured from CT. One set of measurements on a patient's CT (female, 38 years old) is shown in Fig. 9. Data measured from other 20 coronal slices are summarized in TABLE I.

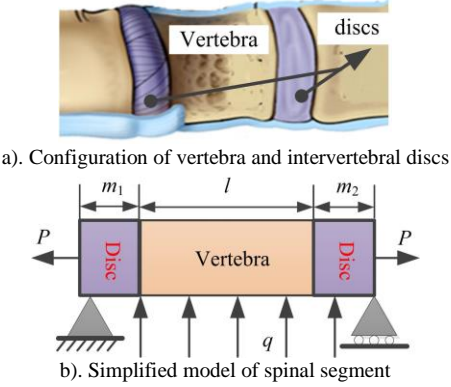

Fig. 8. Actual configuration and simplified model of vertebra segment.

TABLE I
LENGTH OF INTERVERTEBRAL DISC AND T12 SEGMENT

| Types | Upper disc | T12 centrum | Low disc |
|---|---|---|---|
| Length (mm) (M ± SD) | 5.31±0.272 | 23.10±0.445 | 5.67±0.449 |

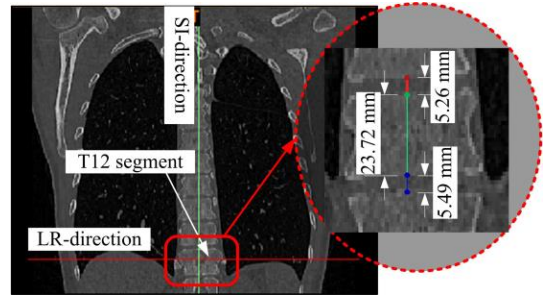

Fig. 9. Posteroanterior radiographs. The image on the right is the zoomed-in image of the T12 segment and two adjoining intervertebral discs.

From TABLE I, the length of the two nearby discs are very close to each other (with a difference of 0.36 mm). Compared with the centrum length, the difference only occupies 1.6%. Thus, it will be convenient to set $m_1 = m_2 = m$ (disc length). In addition, the elastic modulus of bone tissue (18.5 ± 4.9 GPa [37]) is much higher (over 200 times) than the modulus of the disc (75.8 kPa [38]). Thus, the vertebra is seen as a rigid body, and the displacement mainly caused by the deformation of intervertebral discs. Based on the simple beam theory, the differential equation of the deflection curve can be expressed as Eq. (6).

$$\dfrac{d^2 v}{dx^2} = \dfrac{qx^2 - 2R_s x}{2EI} \quad (x \le m) \quad (6)$$

where $v$ is the deflection, $E$ and $I$ are the elastic coefficients, and the inertial moment of the intervertebral discs and $R_S = q(l+2m)/2$ represent the reaction. Then, $v$ and the slope ($\theta$) of the simplified physical model are computed by integral as follows:

$$\begin{cases} \theta = \dfrac{dv}{dx} = \dfrac{q}{6EI} x^3 - \dfrac{q(l+2m)}{4EI} x^2 + c_1 \\ v = \int \theta dx = \dfrac{q}{24EI} x^4 - \dfrac{q(l+2m)}{12EI} x^3 + c_1 x + c_2 \end{cases} \quad (7)$$

where $c_1 = \frac{qm^3}{3EI} + \frac{qlm^2}{4EI}$ and $c_2=0$.

The vertebra displacement in the AP direction ($d_{AP}$) can be represented by the deflection of the beam when $x$ is equal to $m$. Then, by substituting Eq. (4) in to Eq. (7), $d_{AP}$ with respect to $Tv$ is written as Eq. (8)

$$d_{AP} = q_1^{AP} Tv + q_0^{AP} \quad (8)$$

where $q_1^{AP} = \left(\frac{5m^4}{24EI} + \frac{lm^3}{6EI}\right)\frac{p_0 b_w}{V_0}$, $q_0^{AP} = \left(\frac{5m^4}{24EI} + \frac{lm^3}{6EI}\right) p_0 b_w$.

The vertebra displacement in SI direction ($d_{SI}$) is seen as the elongation caused by the axial load ($P$).

$$d_{SI} = \Delta L = \frac{P}{EA/L} = q_1^{SI} Tv + q_0^{SI} \quad (9)$$

where $EA/L$ is the stiffness of discs, $A$ and $L$ are the mean cross-sectional and total length of intervertebral discs, $q_0^{SI}$ is a constant representing the initial state, and $q_1^{SI} = 2p_0 SL/EAV_0$.

## IV. COMPENSATION AND KEY POINT RECOGNITION METHOD

In this section, an active compensation method is proposed, and the key point recognition algorithm based on the thrusting force is described. To compensate for the vertebral motion, the real-time $Tv$ is calculated from the ventilator setting for integrating into clinical routine.

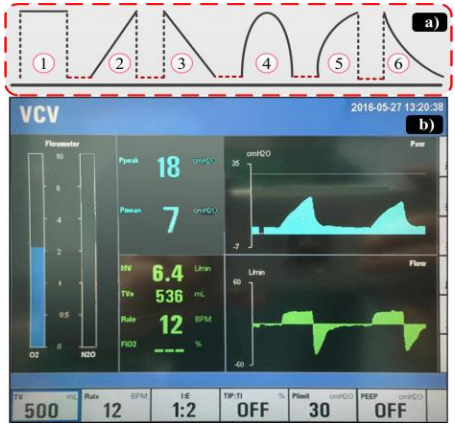

Fig. 10. a) Typical models of ventilator waveforms, which are called Square, Ascenging Ramp, Descenging Ramp, Sine, Exponential Rise, and Exponential Decay, b) Ventilator setting in the operational room.

### A. Model of Tidal Volume

There are 6 basic shapes of waveforms representing the flow velocity used by the ventilator as shown in Fig. 10 a) [39]. Fig. 10 b) is the detailed setting used in spinal surgery. The ventilator works under volume control mode. The square and exponential waveforms were used for inspiratory and expiratory phases, respectively. The other key settings are (1) the ideal maximum tidal volume $Tv_{max}$ = 500 ml, (2) the frequency of respiratory $F_r$ = 12 times/min, and (3) the ratio of the inhale and exhale $rate$ = 1:2. The maximum flow velocity of the exhale phase is about 1.5 times of the constant velocity of the inhale phase. Then, the flow velocity can be written as follows.

$$F(t) = \begin{cases} a & kT \le t < kT + \frac{5}{3} \\ -\frac{3a}{2} b^{t-\frac{5}{3}-kT} + c & kT + \frac{5}{3} \le t < kT + 5 \end{cases} \quad (10)$$

where $F(t)$ represents the flow velocity and the unit is ml/s; $T=5$ s is the period of respiration; $t$ is the time recorded from the starting point of the inhale phase; $a$, $b$, and $c$ are constant coefficients. Since the flow velocity is a continuous variable, the constant coefficients ($a$, $b$, $c$) describing the flow velocity should be constrained by the boundaries and the continuous condition as follows.

$$\begin{cases} \lim_{t \to kT^-} F(t) = 0 \\ \int_{kT}^{kT+5} F(t) dt = 0 \\ \int_{kT}^{kT+\frac{5}{3}} F(t) dt = -500 \end{cases} \quad (11)$$

To optimize the results of the constrained nonlinear equations, the trust-region-dogleg algorithm is used, giving $a = 300$, $b = 0.4602$, $c = 0.0753$. Then $Tv$ is expression of the tidal volume in respect to time can be obtained by integrating the above formula as Eq. (12).

$$Tv(t) = \int_0^t F(t) dt \quad (12)$$

### B. Compensation Method

The real-time displacement of the subject is computed by substituting real-time $Tv(t)$ into the deduced prediction model based on the simplified physical model. Then, the new trajectory of the drilling bit is generated by combining the feed rate together with the predicted displacement based on the finite time control method. Firstly, the displacement is separated by identical short time periods. Since the displacement is very small (less than 5 mm), the target velocity and acceleration for each tiny time period can be seen as a constant. Thus, the trapezoidal velocity profile motion (distance, velocity and acceleration) is generated by empirically setting the acceleration to be 10 times the velocity. With a suitable time period, the continuous displacement motion can be closely approximated by a series of discrete motions. Then the time-related movement segmentation is added to the feed motion performed by RSSSII to compensate for the vertebral motion caused by respiration. In this paper, the tiny time period is 125 ms.

**Algorithm 1**. Active Compensation Algorithm

1: $(Tv_{max}, F_r, rate)$ ← initialize parameters of ventilator
2: $F(t)$ ← calculate flow velocity
3: $Tv(t)$ ← calculate tidal volume
4: $y_{AP}(t)$ ← calculate predicted displacement
5: SPMS and RSSSII **do move**
6: **while** drill bit doesn't reach stop point **do**
7:     $F_{rec}$ and $P_{rec}$ ← recorded from sensors
8:     **if** $\|y_{AP}(t) - P_{rec}\| > H_1$ or $\| F_{rec} \| > H_2$ **then** } Monitor
9:         stop drilling and retreat                     System
10:   **end if**
11: **end while**

The whole compensation process and monitoring system are summarized in Algorithm 1. To ensure patient safety, the real-time position of the drilled bone and drilling force are used as the input data of the monitor system. Based on the experiments, threshold $H_1$ and $H_2$ are set to be 1.2 mm and 10 N.

### C. Feature Extraction of Thrusting Force

Since the drilling bit is invisible during the operation, the thrusting force ($f_i$) is used to recognize the drilling state, which is sensitive to the changes of drilling condition. To remove the digital noise of the sensor, a moving average filter is used and the average force ($\bar{f_i}$) can be calculated by Eq. (13).

$$\bar{f_i} = \frac{1}{n} \sum_{j=i-n+1}^{i} f_j \quad (13)$$

where $n$ is the number of the values used to obtain the average value, which is set as 10 in this paper, $i > n$.

Ideally, the thrusting force recorded in the cortical bone is larger than the force in the cancellous bone. However, the force recorded in the cancellous bone is fluctuated because of the inhomogeneous structure (Fig. 11). The fluctuation will make it more difficult to recognize the drilling states. To restrain the small magnitude force fluctuation, the recognition feature function $A_i$ is computed as Eq. (14).

$$A_i = \begin{cases} D & \bar{f_i}^* > 1 \\ D(\bar{f_i}^*)^3 & 0 \le \bar{f_i}^* \le 1 \\ 0 & \bar{f_i}^* < 0 \end{cases} \quad (14)$$

where $D$ is a constant used to adjust the magnitude of $A_i$ according to the magnitude of $\bar{f_j}$. $\bar{f_i}^*$ is the normalized force according to the $\bar{f_j}$ recorded during outer cortical bone.

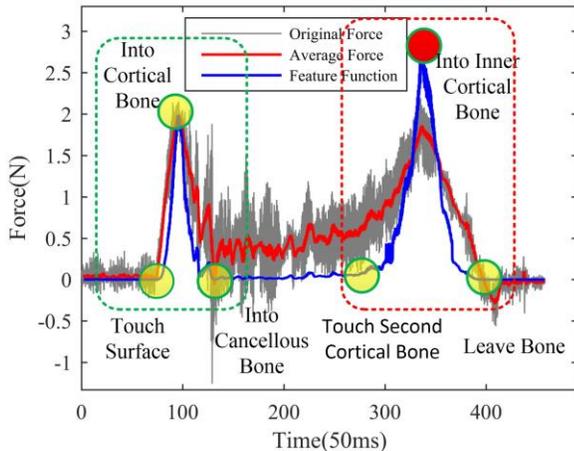

Fig. 11. Drilling force signal. The six dots are the key points of the drilling process, the data in the green rectangle is used to adjust the parameters of the recognition function, and the data in the red rectangle is the key process needed to be detected.

In addition, the thickness of the inner cortical bone may be less than the thickness of the outer cortical bone in reality. This makes it hard to select a dynamic threshold to identify the state in the second cortical bone because the magnitude of feature function may far less than the $\bar{f_j}$ recorded during outer cortical bone. Thus, gain $G$ ($> 1$) is introduced to amplify the high–value part (cortical bone) of the feature recognition curve while the low–value part (cancellous bone) is kept unchanged. This will make it easier to distinguish the transition phase. Thus, the modified feature function ($A_i^*$) is computed as follows:

$$A_i^* = \begin{cases} A_i & A_i \le H_g \\ GA_i & A_i > H_g \end{cases} \quad (15)$$

where $H_g$ is the threshold. The success rate reaches a very high level when $G = 1.2$ and $H_g = 0.5$ based on experiments.

The modified feature function ($A_i^*$) is shown in Fig. 11. Based on $A_i^*$, six key points are defined to help to discriminate the current state of the drilling process and make suitable decisions automatically and timely.

### D. Key Point Recognition Algorithm

The thrusting force of the first layer cortical bone is crucial for the recognition algorithm because it will be used to automatically define the parameters to adapt to the diversity of different patients. The index of the end point of first layer cortical bone $k$ and the force amplitude $D$ for different patients are automatically determined as Algorithm 2.

**Algorithm 2**. Determination of parameter $k$ and $D$
1: $F_{max} \leftarrow 0$
2: **for** $i \leftarrow 1$ to $\infty$ **do**
3:    **if** $\bar{f_i} > F_{max}$ **then**
4:       $F_{max} \leftarrow \bar{f_i}$
5:    **end if**
5:    **if** $\bar{f_i} < K * F_{max}$ **and** $F_{max} > F_{th}$ **then**
6:       $k \leftarrow i;\ D \leftarrow F_{max} - \min(\bar{f_i}),\ (i = 1, 2, \ldots, k)$
7:    **end if**
8: **exit**

Fig. 11 shows that $A_i^*$ decreases toward zero after the drilling bit punctures through the first cortical layer (2nd point). Then, $A_i^{*\,s}$ starts to increase again when the drilling bit arrives at the surface of the inner cortical layer, and the force would stop to increase until the drilling bit punctures the inner cortical bone (5th point). To achieve the desired screw path, RSSSII should stop drilling when the force is located between the 4th and 6th key points in Fig. 11. Then the detailed steps to build the online state recognition algorithm are depicted in Fig. 12. Based on dynamic $D$ and the real-time force, $A_i^*$ is computed. $C_i$ ($i = 1, 2, 3$) is the threshold used to determine the drilling state. Based on experiments, $C_1 = 0.4$, $C_2 = 0.7$, $C_3 = 0.7$.

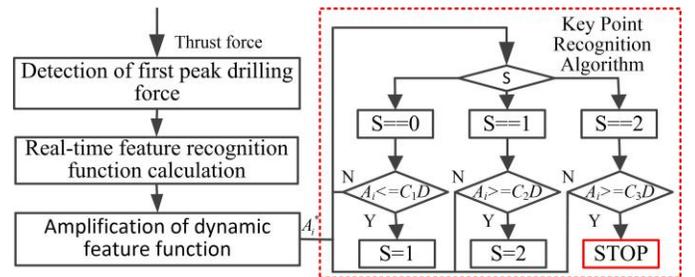

Fig. 12. Key point recognition algorithm.

## V. EXPERIMENT AND DISCUSSION

### A. Experiment Setups

The detailed experimental setups are shown in Fig. 13. The force sensor is attached to the end-effector of RSSSII. The bone drill is mounted on the other side of the force sensor. All experiments were performed on a fresh porcine lamellar bone, which is mounted on the top of the SPMS. The other drilling conditions are set as follows: the diameter of the drill bit = 2.5 mm and the feed rate = 0.5 mm/s.

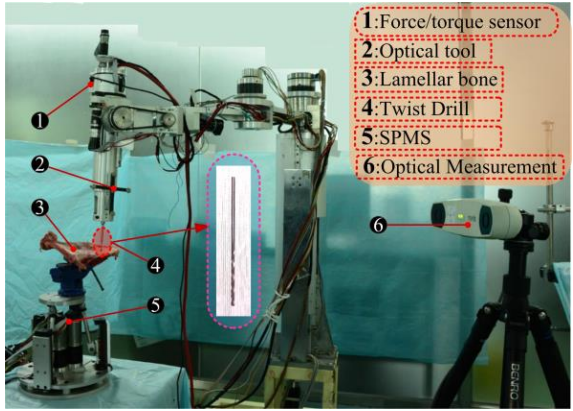

Fig. 13. Setup of the experimental platform.

### B. Parameters Identification Based on PSO Algorithm

The recorded vertebra displacement is proportional to measured $Tv$ as in [35]. The displacement amplitude in the anterior-posterior (AP, about 5 mm) direction is greater than the signals in the superior-inferior (SI, about 2 mm) and left-right (LR, about 1 mm) directions.

Based on the RMSE-DMM, the coif4, bior2.8, and coif5 are finally selected as biases for processing the displacement in AP, SI, and LR direction. The results of the DWT with specified biases are shown in Fig. 14. The distribution of the de-noised AP displacement is close to the raw data, while the de-noised data in LR and SI directions have been compressed. This is because the unavoidable noise from sensors and patients has a larger impact on the signal with a smaller magnitude.

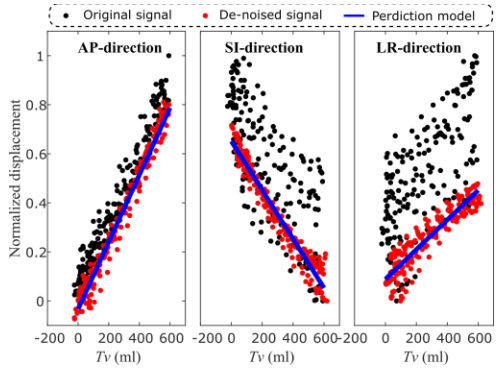

Fig. 14. Results of de-noising data and prediction model in three directions.

The vertebra displacement model (AP and SI directions) with respect to $Tv$ has been described in Section III-B. Since spine configuration is symmetrical in LR direction, the displacement in the LR direction would be ideal zero. However, since the configuration is not perfectly symmetrical and the patient may be tilted to one side on the operation bed, LR displacement will also slightly change according to $Tv$, as shown in Fig. 14. The de-noised displacement in the LR direction ($d_{LR}$) is distributed close to a line. Thus, a linear polynomial is used to describe $d_{LR}$ in this work.

To reflect patient diversity, the parameters of the prediction models are optimized by the PSO algorithm [40]. The PSO is one of the most popular modern heuristic algorithms and it is easy to implement. In this application, since the target is to find the most suitable parameters to make the models adaptive to different patients, the coefficient of determination ($R^2$) is used as the fitness function.

$$R^2 = 1 - \frac{\sum_{i=1}^{n}(g_i - \hat{g}_i)^2}{\sum_{i=1}^{n}(g_i - \bar{g}_i)^2} \quad (16)$$

where $g_i$, $\bar{g}_i$, and $\hat{g}_i$ represent the recorded displacement, average of recorded data, and the average of the prediction displacement of vertebral position in three directions.

The other parameters are set as follows: the population size = 24 and the maximum number of iteration = 2000. The final prediction models are shown in Fig. 14. To further validate the method, the prediction model for different volunteers is calculated by repeating the above-mentioned process and the fitting results for five volunteers are listed in TABLE II.

The average $R^2$ in the LR direction is 81.2%, which is much lower than the average value in AP and SI directions (94.6% and 95%). This is because the LR displacement is mainly caused by small movements, which deviated the weight center of patients. However, based on TABLE II, we conclude that the displacements are well-represented because the largest and the second-largest displacement magnitudes take place in AP and SI directions (approximately 4 mm and 2 mm).

TABLE II
RESULTS OF RECOGNITION ALGORITHM

| $R^2$ | 1 | 2 | 3 | 4 | 5 | Mean |
|---|---|---|---|---|---|---|
| AP | 97% | 94% | 94% | 93% | 95% | 94.6% |
| SI | 94% | 93% | 93% | 95% | 95% | 95.0% |
| LR | 86% | 80% | 75% | 83% | 82% | 81.2% |

### C. Rotation Speed for RAS

The key point detection method is developed based on the trusting force, which is influenced by the spindle speed and the geometry of the drill bit [23]. The tool geometry is determined by doctors according to the size of the vertebra needed to be drilled. As for the experienced surgeons, high speed (over 30000 rpm—revolutions per minute) is popular because it leads to a smoother force, which will be helpful to detect the phase of the drill bit from the cortical bone to cancellous bone or vice versa.

However, a higher spindle speed will decrease the magnitude of the trusting force, making it much easier to be disturbed by environment noise (e.g. digital noise of sensor). This will hinder the RAS to detect the right key point based on the noised force with a small magnitude.

To robustly detect the right stop point for RAS, the suitable spindle speed for RAS is determined by the performance of the experiments with different spindle speeds (12000, 16000, and 20000 rpm). For each spindle speed, three sets of experiments were performed in which Case a) stationary case: drilled bone is stationary; Case b) without compensation case: the tested bone is moving; Case c) with compensation case: the tested bone is moving, while RSSSII has compensated the bone movement. To limit the potential negative influence on the detected result caused by the sub-optimal compensation, only the largest displacement signal in the AP direction was considered. The desired stop point for the drill bit is located inside of the second cortical layer of the lamellar bone to achieve the best biomechanical stability [21]. Thus, the experiment would have failed, if the drill bit stopped before entering the inner cortical bone or after passing the inner cortical layer. The final results of all 96 experiments are shown in Fig. 15.

The success rate of the recognition algorithm for the stationary case is the highest among all three spindle speeds. However, it actually works better with a lower spindle speed (12000 or 16000 rpm). In addition, for the case without compensation and the case with compensation, the success rates decrease when the spindle speed increases, particularly for case c. The success rate dramatically reduced from 90% to 50% when the speed changed from 12000 to 16000 rpm and it further reduced to 40% when speed was 20000 rpm. Thus, 12000 rpm is suitable for the robust detection of the desired stop point for RAS.

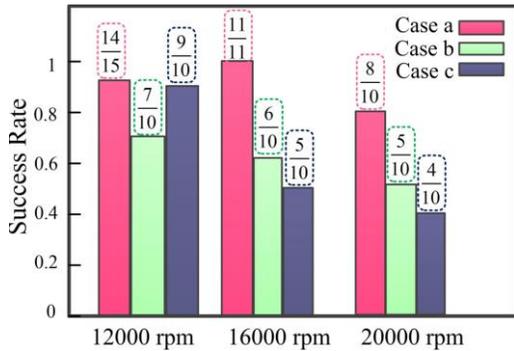

Fig. 15. Success rate of recognition algorithm for Case a, Case b, and Case c with different spindle speeds. stationary case, moving bone without compensation case and with compensation case, respectively.

### D. Validation of the Compensation Method

To validate the performance of compensation, the peaking force when drilling inside the cortical bone was investigated. Based on Fig. 11, the transition phase can be easily recognized when the two peaking forces ($F_{out}$ and $F_{in}$) corresponding to the outer and inner cortical bone layer become larger and more stable. To assess the performance of the compensation method, the three sets of experiments (Case a, Case b and Case c were performed with 12000 rpm spindle speed. The results of $F_{out}$ and $F_{in}$ for all 35 experiments (15 Case a, 10 Case b, and 10 Case c) are shown in Fig. 16.

The range of $F_{out}$ and $F_{in}$ of Case b (without compensation) is wider than the other two cases, meaning that the force profile in Case b is not stable. This is because the relative motion between the drill bit and tested bone is varying in Case b. The different relative motion will lead to a different thrusting force profile with the same feed rate. However, both the range and median/mean $F_{out}$ and $F_{in}$ of Case c (with compensation for moving bone) is close to the results of Case a (stationary case). For the outer cortical layer, the difference between the median $F_{out}$ of Case a and Case b is 1.2 N (38%), while the difference between Case a and Case c is only 0.43 N (13%). For the inner cortical layer, the difference between the median $F_{in}$ are 1.43 N (49.7%) and 0.01 N (0.2%). This means that the compensation method can stabilize the thrusting force to improve the performance of the recognition method.

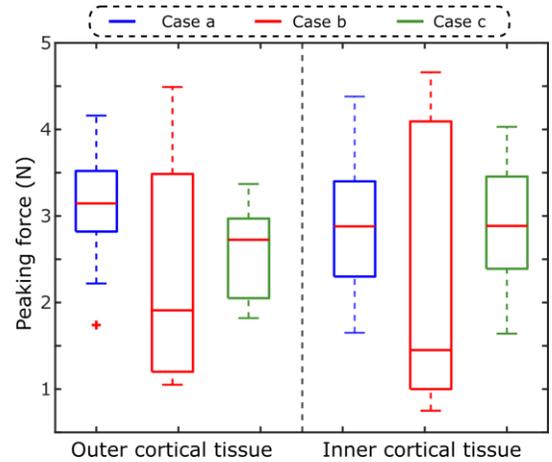

Fig. 16. The peaking force recorded in three sets of experiments. Spindle speed = 12000 rpm.

### E. Validation of the Recognition Algorithm

To validate the effectiveness of the compensation method and recognition algorithm, an additional 60 experiments were carried out with 12000 rpm spindle speed. All experiments were classified into three groups, as in the last section (stationary case, moving bone without compensation and with compensation cases). Three results are selected from different groups (Fig. 17). The raw force was directly measured from the force sensor. To restrain the fluctuation of the sensor, the input force is the mean of each fifty raw data. Then, the average force and feature function can be calculated by Eq. (13) and Eq. (15), respectively.

In Fig. 17 a) and c), the drill bit automatically stopped at the inner cortical bone, however, in Case b, the drill bit stopped at a wrong position located in the outer cortical bone. This is caused by the varying relative motion between the drill bit and the tested bone when bone motion is well compensated. The difference between the thrusting force profile can be featured by two peaking forces ($F_{out}$ and $F_{in}$). For Case b, $F_{out}$ =1.1 and $F_{in}$ =0.9 N are much smaller than $F_{out}$ and $F_{in}$ for the stationary case (3.6 and 3.0 N) and case with compensation (2.9 and 3.7 N) in Fig. 17. In the worst case, unexpected peaking points may appear during the drilling procedure, as shown in Fig. 17 b).

The average thickness of the inner cortical layer of the porcine lamellar bone is 2.11 mm. Thus, the drilling procedure is successful if the residual thickness of the inner cortical bone is ranging (0, 2 mm]. The success rate of the three groups has been listed in TABLE III. Then, three results are randomly selected from the 19 successful cases from the group with compensation (see Fig. 18). Their residual thicknesses are all close to 1.5 mm, demonstrating that the drill bit stopped in the inner cortical bone, and there is a suitable distance between stop point and breakthrough point. This means the modified recognition method works well together with the compensation method.

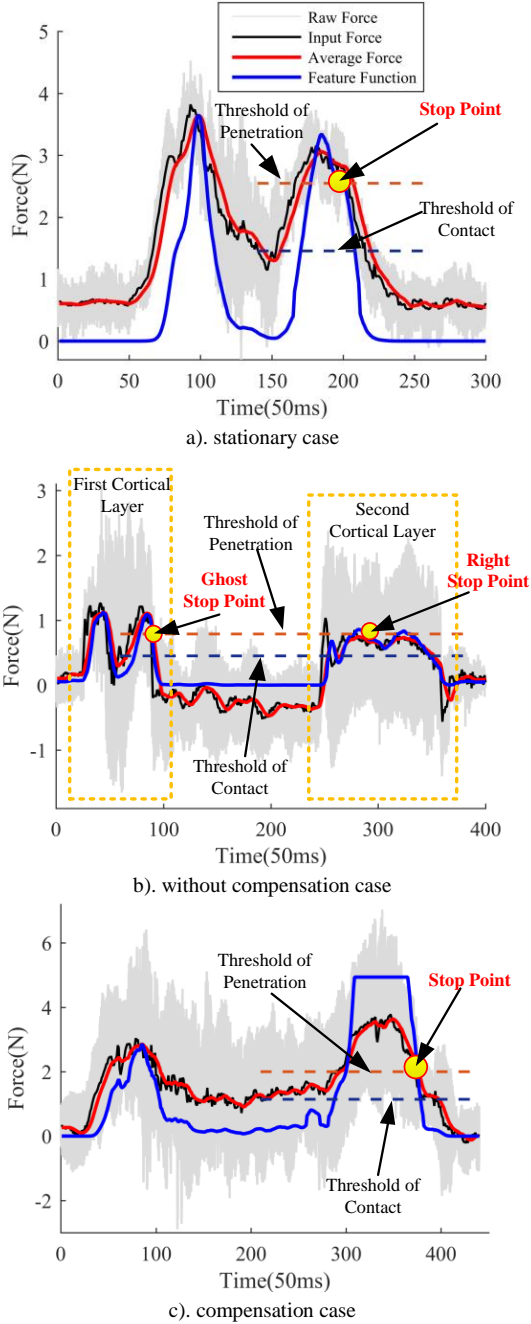

Fig. 17. Stop point detection results. Spindle speed = 12000 rpm.

Compared with previous work [34], the result of the ideal (stationary) case failed once among 20 experiments. This is because the stiffness of the 3-axis linear robot (TMP, Beijing, China) is higher than the 6-DoF serial robotic arm (RSSSII). However, it is noteworthy that the success rate of the compensation has been improved much more than the case without compensation, and it is comparable to the ideal case. This demonstrates that the presented compensation method and recognition algorithm are helpful to automatically detect the right stop point during drilling procedure on the moving vertebra caused by respiration.

TABLE III
RESULTS OF RECOGNITION ALGORITHM

| Case | Ideal Case | Without Compensation | With Compensation |
|---|---|---|---|
| Results | 19/20 | 13/20 | 19/20 |
| Success Rate | 95% | 65% | 95% |

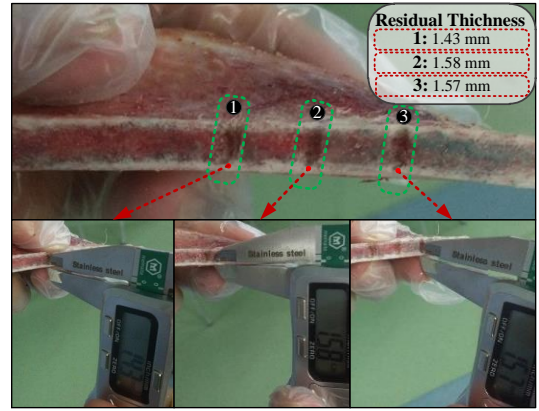

Fig. 18. Residual thickness of inner cortical bone.

## VI. CONCLUSION

This paper presents an automatic robotic-assisted system (RAS) for pedicle drilling procedures and further validation of the proposed state recognition and compensation methods. It paves the way toward clinical practice by taking the unavoidable vertebra displacement caused by respiration into consideration. A method for quickly computing the real-time $T_v$ based on the ventilator's setting was investigated to integrate the proposed pipeline to the current clinical routine. In addition, the spindle speed for RAS is set as 12000 rpm to achieve better recognition performance rather than using the speed preferred by human operators (over 30000 rpm). Finally, the experiments performed on the fresh porcine lamellar bone demonstrate that the proposed method works effectively to automatically identify the suitable stop point for a moving tissue when the compensation method is used. The success rate of the case with compensation is 95%, which is comparable to the stationary case and much higher than the case without compensation (65%). In the future work, for clinical applications, the animal experiments should be carried out, and more realistic factors should be taken into consideration, such as the vertebra rotation and interactive force applied by surgeons.